\documentclass[conference]{IEEEtran}
\usepackage{stmaryrd}
\usepackage{amsfonts}
\usepackage{epic}   
\usepackage{ecltree}    
\usepackage{fancybox}   
\usepackage[latin1]{inputenc} 
\usepackage[dvips]{epsfig}

\usepackage{graphicx,times,amsmath,url} 

\IEEEoverridecommandlockouts    

\begin{document}

\title{\ \\ \LARGE\bf Distributed Evolutionary Computation using REST}

\author{
P.A. Castillo \and M.G. Arenas \and A.M. Mora \and J.L.J. Laredo \and G. Romero \and  V.M Rivas \and J.J. Merelo
\thanks{GeNeura. Department of Architecture and Computer Technology. CITIC. University of Granada. {\tt http://geneura.wordpress.com}  Contact email: {\tt pedro@atc.ugr.es} }}

\maketitle

\begin{abstract}

This paper analises distributed evolutionary computation based on the Representational State Transfer (REST) protocol, which overlays a farming model on evolutionary computation.
An approach to evolutionary distributed optimisation of multilayer perceptrons (MLP) using REST and language Perl has been done. In these experiments, a master-slave based evolutionary algorithm (EA) has been implemented, where slave processes evaluate the costly fitness function (training a MLP to solve a classification problem).
Obtained results show that the parallel version of the developed programs obtains similar or better results using much less time than the sequential version, obtaining a good speedup. 

\end{abstract}

\section{Introduction}
\label{sec:introduction}

Service Oriented Architecture (SOA) \cite{PAPAZOGLOU} is a paradigm for organizing and utilizing distributed computational resources, called services. Using this paradigm, the service providers publish the descriptions (or interfaces) of the services they offer in a service registry, so that the service requesters can discover them and bind to the correspondant service provider.
Web Services are the key point of integration for different applications belonging to different platforms, languages and systems since they are based in a set of standards that make them independent of the underlaying technologies used for providing them.

Although there are several technologies for developing web services (SOAP, REST or XMLRPC among others \cite{compxmlrpc1,compxmlrpc2}), nowadays the main approaches are SOAP (Simple Object Access Protocol) \cite{SOAP,avila01} and REST (Representational State Transfer) \cite{wikiREST}.

SOAP is the traditional, standards-based approach, but the majority of the web services with public API offer REST interfaces, while some of them offer both REST and SOAP and very few offer just SOAP.

All of the major Web Services providers use REST: Twitter, Yahoo's, Flickr, del.icio.us, pubsub, bloglines, technorati, and several others. Both eBay and Amazon have Web Services for both REST and SOAP.

On the other hand, SOAP Web Services are used in lots of enterprise software as well; for example, Google implements their Web Services using SOAP, with the exception of Blogger, which uses XML-RPC, an early and simpler pre-standard of SOAP. 

The philosophies of SOAP and RESTful Web Services are very different. Strictly, SOAP is a protocol for distributed computing, whereas REST adheres much more closely to a web-based design. 
SOAP requires a greater implementation and understanding effort from the client side in difference to REST based APIs, which focuses these efforts on the server side. 

It is important to note that one of the advantages of SOAP is the use of a ''generic'' transport. While REST today uses HTTP/HTTPS, SOAP can use almost any transport to send the request. 
However, one perceived disadvantage is the use of XML because of its verbosity, and the time necessary to parse it.

\medskip

This work continues with our previous research in service oriented
algorithms, as previously stated in \cite{OSGILIATH}, where a
service-oriented platform was presented, or \cite{EVAG}, where studies
about P2P distributed evolutionary algorithms were performed.

In this paper we propose using REST for distributed computation, demonstrating how it could be used for evolutionary computation. 
Our aim is to implement a distributed evolutionary algorithm (EA) using Perl and REST, to solve a costly problem: tuning learning parameters and to set the initial weights and hidden layer size of a multilayer perceptron (MLP), based on an EA and Quick Propagation \cite{FahlmanQP} (QP) to solve classification problems.
This paper continues the research on evolutionary optimisation of MLP (\emph{G-Prop} method) presented in \cite{CastilloNPL,castilloNC}. 
This method leverages the capabilities of two classes of algorithms: the ability of EA to find a solution close to the global optimum, and the ability of the back-propagation algorithm (BP) to tune a solution and reach the nearest local minimum by means of local search from the solution found by the EA.
Instead of using a pre-established topology, the population is initialised with different hidden layer sizes, with some specific operators designed to change them (mutation, multi-point crossover, addition and elimination of hidden units, and QP training applied as operator).
The EA searches and optimises the architecture (number of hidden units), the initial weight setting for that architecture and the learning rate for that net.

The main idea of this paper, which is basically a proof of concept, is to see what are the possibilities of this setup as a meta-computer by implementing an EA using it, and then measuring the speedup when several computers are used at the same time.
The problem we will attempt to solve is a costly classification problem, so that it takes time enough to get some improvement from parallelization. We will only try to measure how running time scales when new (heterogeneous) nodes are added to the system, being the main objective to test if this kind of system is suitable for scientific computation.

\medskip

The rest of this paper is structured as follows:
Section \ref{sec:REST} presents a comprehensive description of REST technology.
Main paradigms of parallel and distributed evolutionary computation are reviewed in Section \ref{sec:pdEA}.
Then, section \ref{sec:method} describes the proposed method, based on a farming model.
Section \ref{sec:results} details the experimental setup and presents obtained results.
Finally, a brief conclusion and future work is presented in section \ref{sec:conclusionsAndFutureWork}.

\section{REST: Representational State Transfer}
\label{sec:REST}

After some years, Internet architects have found an alternative method for building web services in the form of Representational State Transfer (REST)  \cite{wikiREST} .

REST is a style of software architecture for distributed hypermedia systems such as the World Wide Web. The term Representational State Transfer was introduced and defined in 2000 by Roy Fielding in his doctoral dissertation \cite{Fielding2000,Fielding2002}. Fielding is one of the principal authors of the Hypertext Transfer Protocol (HTTP) specification versions 1.0 and 1.1 \cite{wikiREST3,wikiREST4}.

REST-style architectures consist of clients and servers. Clients initiate requests to servers; servers process requests and return appropriate responses. Requests and responses are built around the transfer of representations of resources. A resource can be essentially any coherent and meaningful concept that may be addressed. 

Although REST was initially described in the context of HTTP, is not limited to that protocol. RESTful architectures can be based on other Application Layer protocols if they already provide a rich and uniform vocabulary for applications based on the transfer of meaningful representational state. RESTful applications maximize the use of the pre-existing, well-defined interface and other built-in capabilities provided by the chosen network protocol, and minimize the addition of new application-specific features on top of it.

In a REST environment, clients are not concerned with data storage, which remains internal to each server, so that the portability of client code is improved. 
Servers are not concerned with the user interface or user state, so that servers can be simpler and more scalable. 
Servers and clients may also be replaced and developed independently, as long as the interface is not altered.
Finally, servers are able to temporarily extend or customize the functionality of a client by transferring logic to it that it can execute. 

The client-server communication is further constrained by no client context being stored on the server between requests. Each request from a client contains all of the information necessary to serve the request, and any session state is held in the client. The server can be stateful; this constraint merely requires that server-side state be addressable by URL as a resource. This not only makes servers more visible for monitoring, but also makes them more reliable in the face of partial or network failures as well as further enhancing their scalability.


Main REST web services features are:
\begin{itemize}
\item Simple and lightweight (not a lot of extra XML markup)
\item Human readable format
\item Easy to build (no toolkits required)
\item High performance
\end{itemize}

\section{Parallel and Distributed Evolutionary Algorithms}
\label{sec:pdEA}

We are concentrating on parallel and distributed evolutionary computation applications, which has already  been adapted to several paradigms of parallel and distributed computing (for example, Jini \cite{Jini:Paralelos}, JavaSpaces \cite{Setzkorn2004}, Java with applets \cite{chong99}, MPI \cite{javi:jp2001}, service oriented architectures \cite{myers2003} and P2P \cite{maribel:jp2001}).

There are many ways to implement a distributed EA, one of which is the island model (migration): the population is divided into small subpopulations of the same size assigned to different processors. From time to time each processor selects the best individuals in its subpopulation and it sends them to his nearer processors, receiving as well copies of the best individuals of his neighbours (migration of individuals). All processors replace the worst individuals of their populations. This kind of algorithms is also known as distributed EAs (Tanese \cite{Tanese}, Pettey et al. \cite{Pettey}, Cant\'u-Paz and Goldberg \cite{CantuPazGoldberg}).

Another alternative implementation is global paralelization (\emph{farming}) \cite{FogartyHuang,AbramsonAbela,HauserManner}, in which individual evaluation and/or genetic operator application are parallelized. 
The global model does not divide the population. Instead, such an approach employs the inherent parallelism of evolutionary algorithms (population of individuals). 
The calculations where the whole population is needed (fitness assignment and selection) are performed by the master and all remaining calculations which are performed for one or two individuals can be distributed to a number of slaves. 
The slaves can perform recombination, mutation and the evaluation of the objective function separately (these calculations can be done in parallel). This is known as synchronous master-slave structure.
A nearly linear speedup of the calculation time may be achieved (as long as the evaluation time of the objective function is higher than the communication time between master and slaves).
The global model is a simple way (and inherent to every evolutionary algorithm) to reduce very long computation times. 


Although many approaches to distributed EAs \cite{Goldberg} can be found in bibliography, in this paper we do not intend to innovate in that sense, but in the implementation (because implementation matters \cite{jjiwann2011}).

\section{Master-Slave based EA implementation using REST and Perl}
\label{sec:method}

An ideal client-server implementation of a distributed EA could be a server process with several threads. Each thread would include a population, and would communicate with other threads through the shared code among them. Each thread would use an own tail of individuals to send to other threads. Each thread would evaluate its individuals in different remote computers, carrying out the communication using a REST server.

\begin{figure}[!ht]
\begin{center}
\epsfig{file=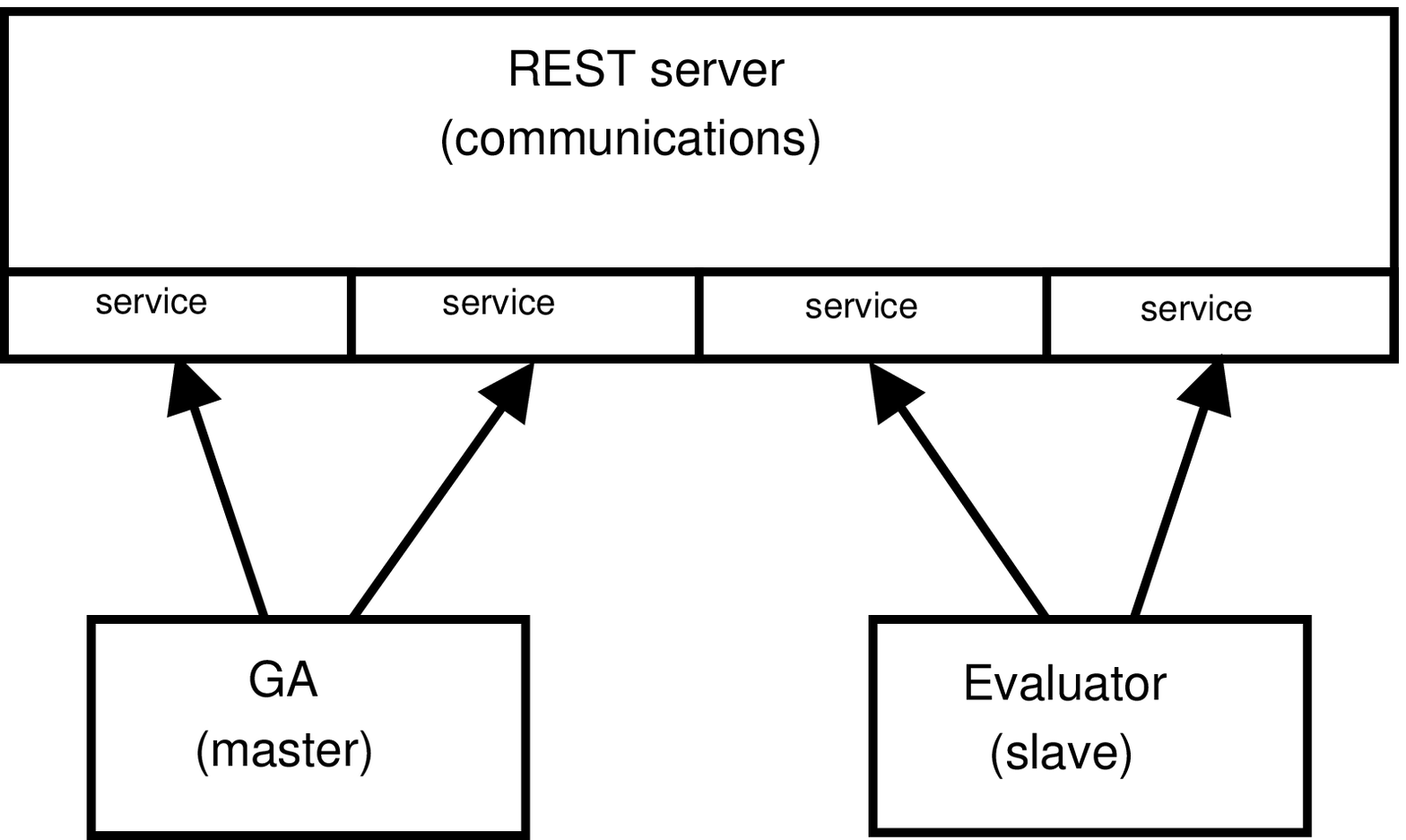,width=7cm}
\caption{Schema of the master-slave based GA implemented in the second experiment. The master process runs the GA and the slave processes evaluate the fitness function.} 
\label{fig:esquema}
\end{center}
\end{figure}

\begin{figure*}[!ht]
\begin{center}
\setlength{\tabcolsep}{2mm}
\renewcommand{\arraystretch}{1.2}
\begin{tabular}{|cc|cc|}

\hline

\begin{tabular}{l}
 use Dancer; \\
 my \$src = ""; \\ 
 get '/' =$>$ sub \{ \\
 \qquad return "Hello World"; \\
 \}; \\
 get '/uploadcode/:code' =$>$ sub \{ \\
 \qquad \$src = params-$>$\{code\}; \\
 \qquad return "ok"; \\
 \}; \\
 get '/downloadcode/' =$>$ sub \{ \\
 \qquad return \$src; \\
 \}; \\
 Dancer-$>$dance; \\
\end{tabular}

& & &

\begin{tabular}{l}
 use LWP; \\
 \$c = new LWP::UserAgent; \\ 
 \$c-$>$agent("RESTzilla"); \\
 \$r = new HTTP::Request GET =$>$  \\
 \qquad 'http://127.0.0.1:3000/downloadcode/'; \\
 \$u = \$c-$>$request(\$r); \\
 \# shows the fitness function received \\
 print \$u-$>$content; \\
 \# evaluate the received function \\
 eval( \$u-$>$content ); \\
\end{tabular}

\\ 

\hline

\end{tabular}
\caption{REST programming example: server (left) and client (right). In this example the REST server deploys three services, while the client first obtains the fitness function as Perl source code by calling the corresponding service, and then evaluates that function.}
\label{fig:ejemploREST}
\end{center}
\end{figure*}

However, as we cannot use a threaded version of the Perl modules, our implementation will focus on the most time consuming operation in G-Prop: the fitness function evaluation.
The whole evolutionary algorithm is run on the master and only the objective function is sent to the slaves for evaluation (as shown in Figure \ref{fig:esquema}). 

The whole system can be sketched as follows:
\begin{enumerate}
  \item The EA process sends the fitness function code to the REST server and creates the EA population.
  \item Some clients connect the REST server and load the fitness function sent as Perl code from the server (Figure \ref{fig:ejemploREST} shows an example of server and client processes implementations to upload and download the fitness function source code).
  \item The EA process sends non-evaluated individuals to the server.
  \item The clients ask for individuals to the server in order to evaluate them.
  \item The clients evaluate individuals and send the result back to the server.
  \item The EA process obtains evaluated individuals from the server and continues the evolutionary loop.
  \item The EA terminates after a fixed number of generations (it sends a termination message throughout the REST server to the clients that remain ready to attend new workloads).
\end{enumerate}

The server in these experiments is mainly used for scheduling and balancing the tasks among the different clients; the network itself is used for communication, but all the interchange of information among clients must be cleared by the central server. However, one of the objectives of the work presented in this paper has been to create an infrastructure that would get rid of the bottleneck represented by the central server in these experiments.

\bigskip

Implementation was carried out using the {\tt Perl Dancer} module \cite{perldancer2011,perldancer2011b} for the Perl programming language, for its stability and the familiarity of the authors with this language  \cite{perl-ea,optimizing-meta,jjiwann2011}. In addition, servers are easy to implement and deploy.

As an example, Figure \ref{fig:ejemploREST} shows the source code of a REST server that deploys three services, and a client that obtains the fitness function (as Perl source code) by calling the corresponding service (and then evaluates that function).

The evolutionary algorithm has been implemented using the Algorithm::Evolutionary (A::E) library \cite{perl-ea,jjSOCO2010}. Version 0.76.2 is used in this work, available at {\sf http://opeal.sourceforge.net } under GPL license.

The full source code (servers, GA and evaluators) and experiment data are available under GPL at: \\  
{\sf http://atc.ugr.es/pedro/GProp-REST.tgz}

In this work, we adapt \emph{G-Prop} as a distributed EA using REST following the detailed structure.
\emph{G-Prop} method has been fully described and analysed out in previous papers (see \cite{CastilloNPL,castilloNC}), thus we refer to these papers for further details.
In most cases, evolved MLP should be coded into chromosomes to be handled by the genetic operators, however, G-Prop uses no binary codification, instead, the initial parameters of the network are evolved using specific variation operators such as mutation, multi-point crossover, addition and elimination of hidden units, and QP training applied as operator to the individuals of the population.
The EA optimises the classification ability of the MLP, and at the same time it searches for the number of hidden units (architecture), the initial weight setting and the learning rate for that net.

Only ``default'' parameters have been used (genetic operators were applied using the same application rate).
No parameter tuning has been done, since we do not intend to find the optimal ones, but to prove feasibility of the implementation.

\section{Experimental setup and results} 
\label{sec:results}

The tests used to assess the accuracy of a method must be chosen carefully, because some of them (toy problems) are not suitable for certain capacities of the BP algorithm, such as generalization \cite{FahlmanBENCHMARKS}. Our opinion, along with Prechelt \cite{Prechelt94c}, is that, in order to test an algorithm, real world problems should be used.

\subsection{The ''Glass'' Classification Problem}
This problem consists of the classification of glass types, and is also taken from \cite{Prechelt94c}. The results of a chemical analysis of glass splinters (percent content of 8 different elements) plus the refractive index are used to classify the sample to be either float processed or non float processed building windows, vehicle windows, containers, tableware, or head lamps. This task is motivated by forensic needs in criminal investigation.
This dataset was created based on the glass problem dataset from the UCI repository of machine learning databases.
The data set contains 214 instances. Each sample has 9 attributes plus the class attribute: refractive index, sodium, magnesium, aluminium, silicon, potassium, calcium, barium, iron, and the class attribute (type of glass).

The main data set was divided into three disjoint parts, for training, validating and testing. In order to obtain the fitness of an individual, the MLP (in the slave processes) is trained with the training set and its fitness is established from the classification error with the validating set. 
Once the EA (in the master process) is finished, when it reaches the limit of generations, the classification error with the testing set is calculated: this is the result shown in tables.

Up to 4 computers have been used to run the algorithm and to obtain results both in sequential and parallel versions of the program. Experiments were conducted running the server process on a Ubuntu/Linux machine, while the clients were run on a Windows 7 with the Cygwin\footnote{http://www.cygwin.com} environment and on Ubuntu/Linux machines.
Computer speeds range from 1.5 Ghz to 2 Ghz and are connected using the ethernet network of the university (with a high communication latency, i.e. an average \emph{ping} of 7 ms). 
No experiments using homogeneous computer network have been done, because our aim is to demonstrate potential of distributed EA using web services.

As stated before, the EA was executed using the ''default'' parameter values (shown in Table \ref{table:parametros}).

\begin{table}[!h]
\begin{center}
\caption{\scriptsize{List of parameters used to execute the EA.}}
\label{table:parametros}
\begin{tabular}{|c|c|}
\hline 
Parameter & Value \\
\hline
\hline
number of generations & 100 \\
\hline
individuals in the population & 100 \\
\hline
$\%$ of the population replaced & $30\%$ \\
\hline
number of hidden units & ranging from 2 to 90 \\
\hline
epochs to calculate fitness & 300 \\
\hline
\end{tabular}
\end{center}
\end{table}


\subsection{Obtained Results}

Time was measured using the ''gettimeofday'' function in order to achieve a good precision.
Time taken to run the EA is reported in Table \ref{tabla:resultados}.
Sequential version of the program was run in the faster machine; and in parallel runs, 
the EA (master process) was run on the faster machine while the evaluators were run on slower machines.
In this experiment we are not interested on comparing results against other authors, but in using a costly problem that justifies using a farming model.

Results obtained can be shown in Table \ref{tabla:resultados}.

\begin{table}[!h]
\begin{center}
\caption{\scriptsize{Results (error $\%$ and time) obtained using both the sequential and the parallel versions (up to 4 evaluators-slaves are used in the farming model). Comparable classification ability is obtained, while time is improved as the number of evaluators is increased.}}
\label{tabla:resultados}
\begin{tabular}{|c|c|c|c|}
\hline 
Model         &  Error ($\%$)  &  Time (seconds)  \\
\hline
\hline
Sequential    &   33 $\pm$ 2   &    1215 $\pm$ 104   \\
\hline
\hline
\multicolumn{3}{|l|}{Master-slave} \\
\hline
\ \ \ 1 eval. &   33 $\pm$ 3   &    1308 $\pm$ 114  \\
\hline
\ \ \ 2 eval. &   32 $\pm$ 3   &    719 $\pm$ 96   \\
\hline
\ \ \ 3 eval. &   32 $\pm$ 2   &    522 $\pm$ 87   \\
\hline
\ \ \ 4 eval. &   32 $\pm$ 3   &    424 $\pm$ 92   \\
\hline
\hline
\end{tabular}
\end{center}
\end{table}

Classification errors show a comparable algorithmic result.
However, better results in time are obtained parallelizing the problem between several computers.

\begin{figure}[!ht]
\begin{center}
  \epsfig{file=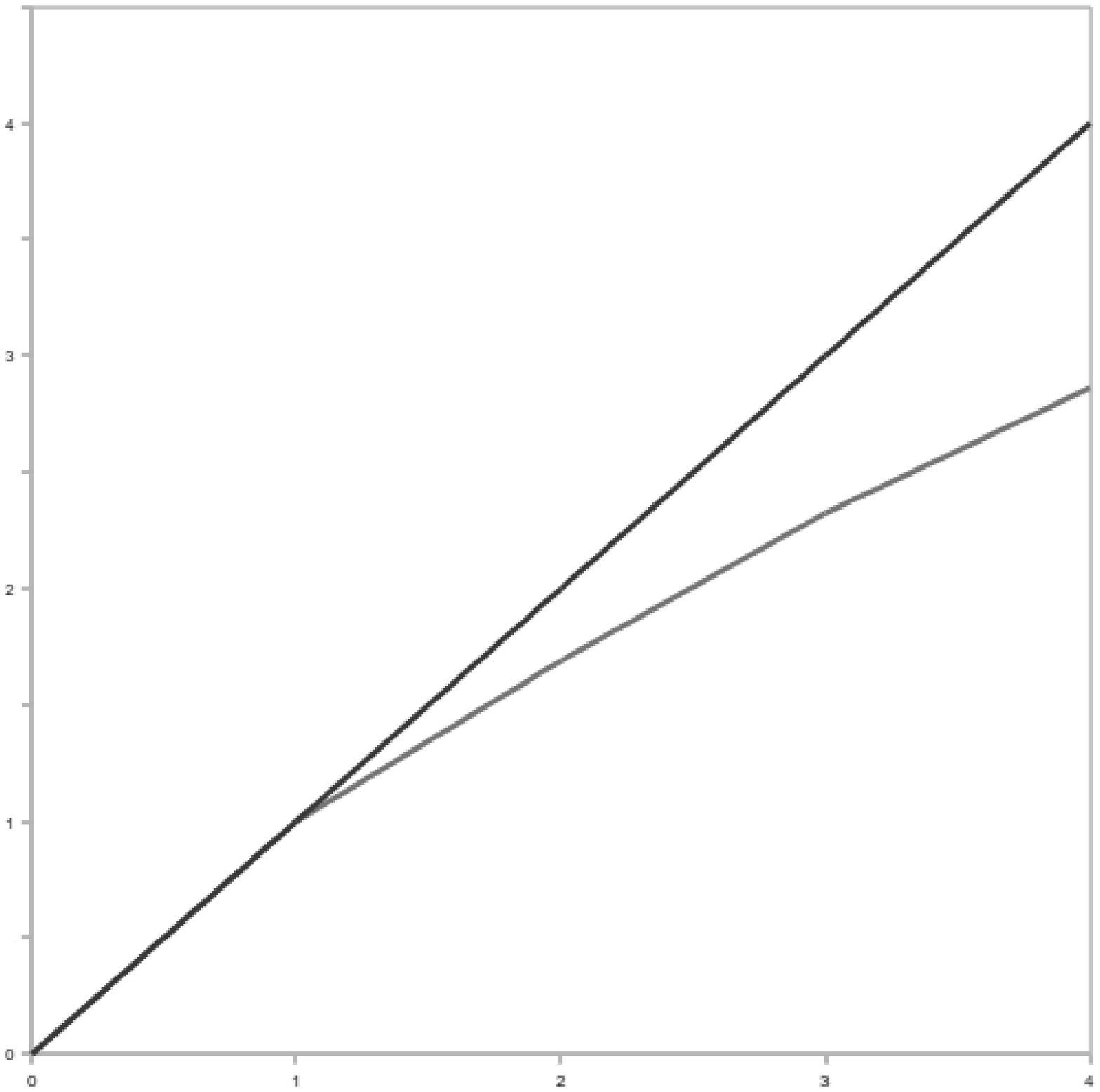,width=7cm}
\caption{Plot of the speedup (dashed line) and $f(x) = x$ function (solid line). Although speedup is not lineal, it can be seen that simulation time is improved using several computers as clients dedicated to evaluate the inidvidual fitness functions increase.}
\label{fig:ganancia}
\end{center}
\end{figure}

Figure \ref{fig:ganancia} shows that speedup does not equals the number of computers used; however, simulation time is improved using several computers. Thus, as adding new evaluators (heterogeneous computers running a Perl process) is an easy and costless task, we could take advantage of this system structure to solve costly optimization problems.
Moreover, results could be better if a dedicated communication network was used, however, the university ethernet network is overloaded and that implies a high latency in communications between processes.

\section{Conclusions and Work in Progress}
\label{sec:conclusionsAndFutureWork}

This paper presents a new parallel-distributed computation implementation using REST and web services that shows the useful this new technology can be in the field of evolutionary computation.

To implement and use communications using REST it is not necessary running virtual machines (as in Java programming), nor daemons, just only to install several libraries available for almost any programming language.
Moreover, an arbitrary number of computers (clients-evaluators) can be added to the system, making it more efficient.

In these experiments, we have demonstrated that REST can be used as communication protocol for distributed evolutionary computation, obtaining a good speedup.
Results could improve using a dedicated communication network instead of the overloaded network of the university.

REST provides a common interface that can be called from almost any programming language. 
Thus, programs can be written in any language and can share data without the need of worrying about the message formats or communication protocols.

At the same time, it does not overload too much the network.
Using other distributed systems, such as Jini \cite{jiniFAQ,Jini:FEA2000}, the network traffic is so high that when a high number of computers are used, communication becomes difficult.

\medskip

A future in which different remote computers offer services to the scientific community can be imagined: for example, all the services available at the moment by means of HTML forms could be implemented easily as services.

As future research, it is very important adding support for REST to existing distributed EA libraries in order to allow the implementation of multi-language EAs.
Another possibility is to test P2P architectures, where each computer communicates only with one or two computers in the network. 
It would be very interesting to parallelize the proposed method using random topologies, in such a way that a ''servent'' (server/client) can enter or leave the network at any moment.


\section*{Acknowledgements}

This work has been supported in part by 
the CEI BioTIC GENIL (CEB09-0010) MICINN CEI Program (PYR-2010-13) project, 
the Junta de Andaluc\'{\i}a TIC-3903 and P08-TIC-03928 projects, and 
the Ja\'en University UJA-08-16-30 project.

\end{document}